\documentclass[runningheads]{llncs}
\usepackage{graphicx}
\usepackage{hyperref}

\begin{document}
\title{Deep Point-wise Prediction for Action Temporal Proposal}
%
%
%
\author{Luxuan Li$^{*\dag}$\and
Tao Kong$^{*\ddag}$ \and
Fuchun Sun$^{\dag}$ \and
Huaping Liu$^{\dag}$}
\authorrunning{Luxuan Li et al.}
%
\institute{$^{\dag}$Department of Computer Science and Technology, Beijing National Research Center for Information Science and Technology (BNRist), Tsinghua University\\
$^{\dag}$ByteDance AI Lab\\
\email{\{llx17@mails, fcsun, hpliu\}@tsinghua.edu.cn}\\
\email{taokongcn@gmail.com}
\footnotetext{*The first two authors contribute equally to the paper.}}
\maketitle              
\begin{abstract}
Detecting actions in videos is an important yet challenging task. Previous works usually utilize (a) sliding window paradigms, or (b) per-frame action scoring and grouping to enumerate the possible temporal locations. Their performances are also limited to the designs of sliding windows or grouping strategies. In this paper, we present a simple and effective method for temporal action proposal generation, named Deep Point-wise Prediction (DPP). DPP simultaneously predicts the action existing possibility and the corresponding temporal locations, without the utilization of any handcrafted sliding window or grouping. The whole system is end-to-end trained with joint loss of temporal action proposal classification and location prediction.

We conduct extensive experiments to verify its effectiveness, generality and robustness  on standard THUMOS14 dataset. DPP runs more than 1000 frames per second,  which largely satisfies the real-time requirement. The code is available at \href{https://github.com/liluxuan1997/DPP}{https://github.com/liluxuan1997/DPP}.

\end{abstract}
\section{Introduction}

\begin{figure}
 \centerline{\includegraphics[width=0.8\textwidth]{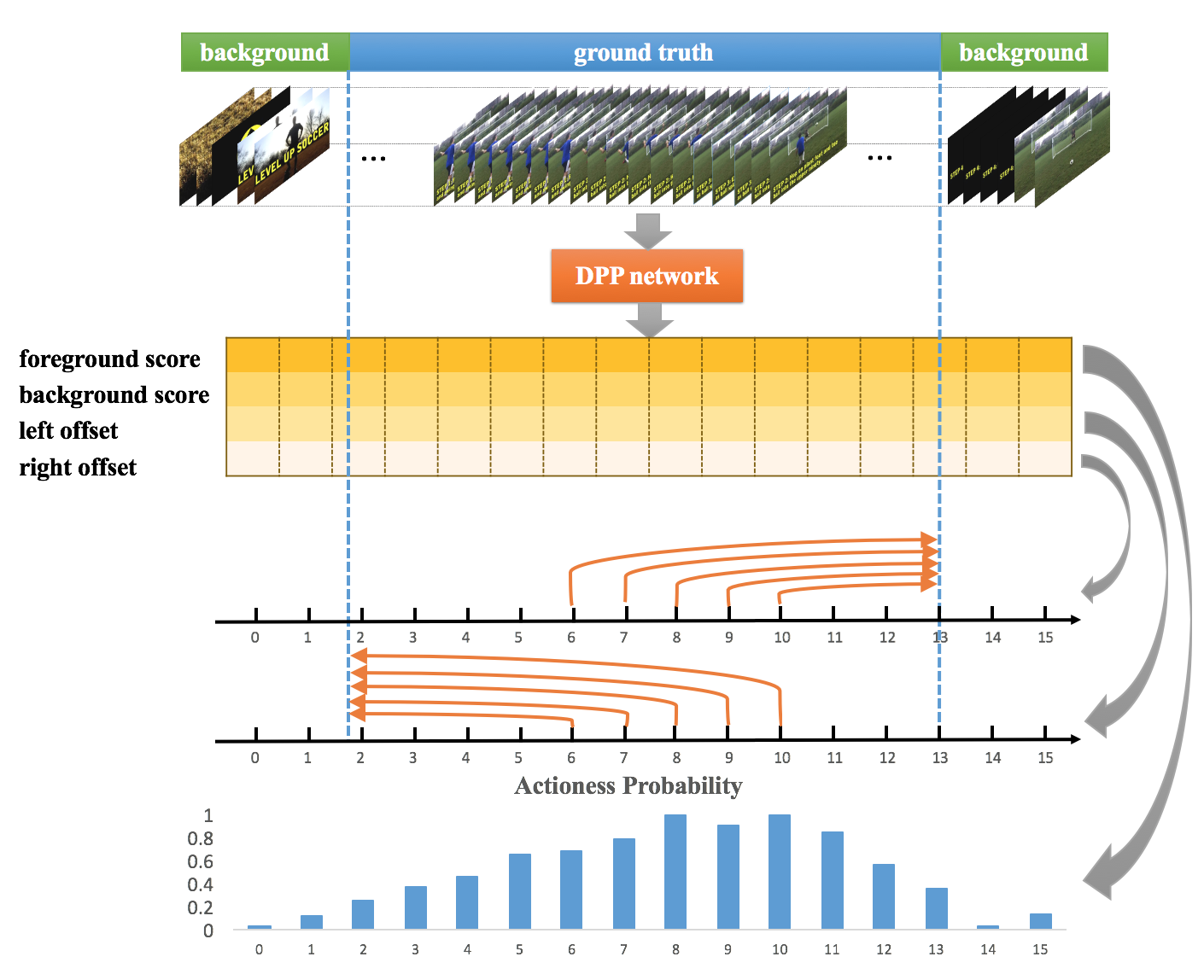}}
 \caption{Overview of DPP. For time points in the sequence of an untrimmed video, DPP directly predicts the probability of action existence and the corresponding starting and ending offsets.}
 \label{fig:overview}
\end{figure}

Despite huge success in understanding a single image, understanding videos still needs further more exploration. Temporal action proposal generation, which aims to extract temporal intervals that may contain an action, has drawn lots of attention recently. It is a challenging task since high quality proposals not only require accurate classification of an action, but also require precise starting time and ending time. 

Previous temporal action proposal generation methods can be generally classified into two main types. The first type is to generate proposals by sliding windows. These methods first predefine a series of temporal windows with fixed lengths as proposal candidates. Then those proposal candidates are scored to indicate the probability of action existence. Finally ranking is applied to get top proposals. Early works like SST \cite{sst} and SCNN-prop \cite{shou2016temporal} try to get high recall by generating dense proposal candidates. SST generates $k$ proposals at each time step by utilizing RNN. TURN \cite{turn} and S3D \cite{s3d} add boundary regression network to get more precise starting and ending time. However, the disadvantages of the sliding window methods are obvious: (1) High-density sliding windows cause great cost of time; (2) Without boundary regression network, the temporal boundaries are not so precise; (3) Sliding windows require multiple predefined lengths and strides, thus introducing additional hyper-parameters of design choices.

The second type is to generate proposals by actioness grouping. These methods evaluate the probability of action existence for each temporal point and group points with high actioness scores to form final proposals. For example, TAG \cite{ssn} first uses an actioness classifier to evaluate the actioness probabilities of individual snippets and generates proposals by classic watershed algorithm \cite{watershed}. BSN \cite{bsn} adopts three binary classifiers to evaluate starting, ending and actioness probabilities of each snippet separately. Then it combines all candidate starting and ending locations as proposals when the gaps between them are not too far. Methods based on actioness score tend to generate more precise boundaries. However, quality of proposals generated by this type of methods highly depends on the grouping strategy. Besides, evaluating actioness probabilities for all points and grouping them limit the processing efficiency.

How we humankind recognize and localize a video action? Do we need pre-defined windows and scanning the whole video sequence? The answer is obviously no. For any single frame in a video, human can directly distinguish if an action happens. And sometimes, human even do not need to see the very start or end of the action but can predict the location.

Inspired by this, we present a simple yet effective system named Deep Point-wise Prediction Network (DPP) to generate temporal action proposals. Our method can be divided into two sibling streams: (1) predicting action existing probability for each temporal point in feature maps; (2) predicting starting time and ending time respectively for each position that potentially contains an action.  The whole architecture consists of three parts. The first part is backbone network to extract high level spatio-temporal features. The second part is Temporal Feature Pyramid Network (TFPN), which is inspired by  Feature Pyramid Network (FPN) \cite{lin2017feature} for object detection task. The third part includes a binary classifier for actioness score and a predictor for starting and ending time. The whole system is end-to-end trained with joint loss of classification and localization.

In summary, the main contributions of our work are three-fold:

\begin{itemize}
    \item We propose a novel method named Deep Point-wise Prediction  for temporal action proposal generation, which can generate high quality temporal action proposals with precise boundaries in real time.
    \item Our proposed DPP breaks through the performance limitation of sliding window based methods. It needs no extra design for predefined sliding windows or anchors. Also, with different backbone networks, DPP gets promising results.
    \item We evaluate DPP on standard THUMOS 2014 dataset, and achieve state-of-the-art performance.
\end{itemize}

\section{Related Work}
\textbf{Action Recognition.} Action Recognition is an important task of video understanding. Architectures of this task always consist of two part: spatio-temporal feature extraction network and category classifier. Since action recognition and temporal action proposal generation both need spatio-temporal features for the following steps, this task is worthy of investigation. Earlier works like improved Dense Trajectory (iDT)\cite{wang2013action} use traditional feature extraction method consists of HOF, HOG, and MBH. With the development of convolutional neural network, many researchers adopt two-stream network\cite{feichtenhofer2016convolutional}for this task. It combines 2D convolutional neural network and optical flow to capture appearance and motion features respectively. Recently, as kinds of 3D convolutional neural networks such as C3D\cite{tran2015learning}, P3D\cite{qiu2017learning}, I3D\cite{carreira2017quo} and 3D-ResNet\cite{hara2018can} appear, adopting 3D convolutional neural network to extract spatio-temporal feature is getting more and more popular\cite{sst,carreira2017quo,s3d,chao2018rethinking}.

\noindent\textbf{Temporal Action Proposals and Detection.} Since natural videos are always long and untrimmed, temporal action proposals and detection have aroused intensive interest from researchers\cite{turn,ssn,sst,s3d,chao2018rethinking,hara2017learning}.  DAP\cite{escorcia2016daps} leverages LSTM to encode the video sequence for temporal features. SST\cite{sst} presents a method combined C3D and GRU to generate temporal action proposals, trying to capture long-time dependency. SCNN-prop\cite{shou2016temporal} adopts multi-scale sliding windows to generate segment proposals. Then it uses 3D convolution neural network and fully-connected layers to extract features and classify proposals separately. Recent studies focus more on how to get proposals with precise boundaries. TURN\cite{turn} applies a coordinate regression network to adjust proposal boundaries. CBR\cite{gao2017cascaded} proposes cascaded boundary regression for further boundary refinement. Other methods like TAL-net\cite{chao2018rethinking} modifies Faster-RCNN to fit temporal action proposal generation task.

For temporal action detection, methods can be divided into two main types: one-stage\cite{ssn,lin2017single,gao2017cascaded,s3d,shou2016temporal} and two-stage\cite{sst,turn,bsn}. One-stage methods like S3D\cite{s3d} generate temporal action proposals and make classification simultaneously. While two-stage methods such as TURN\cite{turn} and BSN\cite{bsn} generate proposals first and re-extract features to classify those proposals.

\section{Approach}
In this section, we introduce the proposed Deep Point-wise Prediction Network and how it works in details. 

\begin{figure*}
 \centering
 \includegraphics[width=\linewidth]{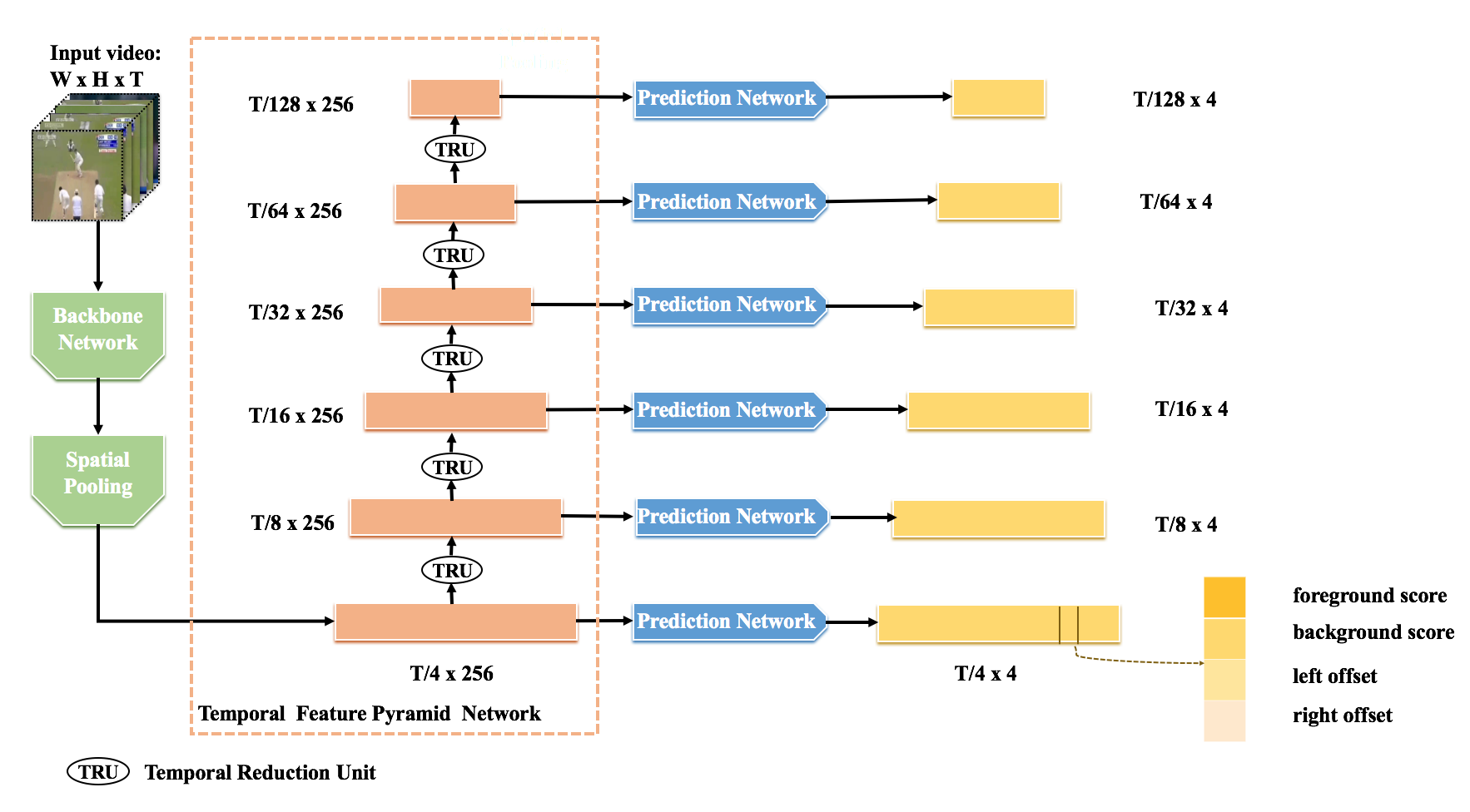}
 \caption{The architecture of our Deep Point-wise Prediction Network.}
 \label{arch}
\end{figure*}

\subsection{Deep Point-wise Prediction Network}
\label{arch_txt}

As shown in Figure \ref{arch}, Deep Point-wise Prediction Network consists of three sub-networks, which are backbone network, Temporal Feature Pyramid Network, and prediction network.

\textbf{Backbone Network.}
We use backbone network and spatial pooling to generate the first-level feature map from a video sequence\footnote{We contrast different backbones in our experiments.}.  More specifically, given a video sequence with shape of $T\times H\times W \times 3$, through backbone network, we get a feature map with shape of $\frac{T}{8} \times \frac{H}{16} \times \frac{W}{16} \times C$, where $T$ is the frame number, $H$ and $W$ are height and width respectively, $C$ is output channel varying with backbone networks. Then we adopt a transpose 3D convolutional layer to upsample the feature map in $T$ dimension and a 2D average pooling layer to pool the spatial features. Finally, we get our first-level temporal feature map with the shape of $\frac{T}{4} \times 256$.

\textbf{Temporal Feature Pyramid Network.}
The core unit of Temporal Feature Pyramid Network is the Temporal Reduction Unit. It receives current feature map as input and outputs next feature map with larger receptive field in each point. And it consists of four 1D temporal convolutional layers with the first three layers of stride 1 and last layer of stride 2. As a result, every feature map is half size of last feature map in temporal dimension. TRU between different levels share the same weights.

\textbf{Prediction Network.}
Prediction Network is applied on different feature maps and generates predictions for every point. The first part is a binary classifier to generate foreground and background score. The second part is a predictor to generate left offset and right offset of proposals. Both parts are achieved by 1D convolutional operation.

\subsection{Label Assignment}
\label{ms}
During training, we need to assign actioness label to every output point according to the ground truth. We design a simple but effective label assignment strategy here. First, points in feature maps are mapped into time points in the original video. For example, for a point in $l_{th}$-level feature map with position $t = \{0, 1, \cdots, T_{l}\}$, its corresponding position in the original video is $2^{l+1}(t+0.5)$. If the corresponding position of a point is inside any ground truth, we define it as a positive point. Further restriction for positive labels is introduced in Section \ref{sa}. Since there is no overlap in adjacent ground truths, a point can only be inside one ground truth. While previous methods whether sliding window based or actioness grouping based adopt a temporal Intersection over Union (tIOU) threshold strategy to define positive proposals and assign corresponding ground truth proposals\cite{s3d,gao2017cascaded,turn,sst,ssn,escorcia2016daps}. Their predefined segments may have overlap with more than one ground truths simultaneously. Compared with the tIoU based matching strategy, our label assignment process is more simple and straightforward.

\subsection{Scale Assignment}
\label{sa}
To predict the proposal location for every point, we try to learn transformation of left offset and right offset between ground truths and current point. Specifically, for points in $l_{th}$-level feature map with position $t = \{0, 1, \cdots, T_{l}\}$ and corresponding ground truth proposal with boundary $(t_{start}, t_{end})$, our localization target is:
\begin{equation}
\label{reg_target}
s_{1} = \lambda \log \frac{2^{l+1}(t+0.5)-t_{start}}{2^{l+1}}, \\
s_{2} = \lambda \log \frac{t_{end}-2^{l+1}(t+0.5)}{2^{l+1}}
\end{equation}
where $l$ indicates that the point is from $l_{th}$ feature map, $T_l$ is the length of this feature map, $2^{l+1}(t+0.5)$ projects the point in feature map into the original input video sequence. $\lambda$ is a coefficient which is set as 3.0 in our training to control the importance of localization part in final loss.

As we can learn from label assignment strategy in section \ref{ms}, a ground truth may be assigned to different points in different level feature maps. And if we keep all these positive points for training, it can be difficult with large scale variations in boundary offsets. Also, as a result of fixed sizes of convolutional kernels, receptive fields of points in the same level feature map are same and points in higher level feature map tend to have bigger receptive fields. And it is hard for a point to predict proposal boundaries far from its receptive field. In $l_{th}$ feature map, the stride of adjacent points is $2^{l+1}$. And its receptive field size is several times as the stride. Here, we want to restrict target left offset and right offset around receptive field of current point. So we divide the original localization targets by default stride of corresponding feature maps to regularize them. For target offsets close to default stride of corresponding feature maps, this operation centers them around 1. And the log function further centers them around 0. We add additional restrictions for positive points as below:
\begin{equation}\label{eq:strict_eta}
s_1,s_2 \in [-\eta, \eta]
\end{equation}
where $\eta$ is a parameter to control the localization range. 
Note that points regarded as positive in Section \ref{ms} but do not satisfy condition in this Eq.\ref{eq:strict_eta} will be ignored during training. As $\eta$ increases, a ground truth is likely to be optimized by more feature maps.

In conclusion, Eq.(\ref{reg_target}) computes the regularized left offset and right offset between each time point in feature maps and corresponding ground truth proposals. With predictions from our regressor, we can easily get the final boundaries by inverse transformation of Eq.(\ref{reg_target}). Eq.(\ref{eq:strict_eta}) selects valuable boundary prediction targets for training.

\subsection{Loss Function}
Our loss consists of two parts which are action loss and localization loss respectively. The overall loss is combination of above two loss defined as:
\begin{equation}
L = L_{act} + L_{loc}.
\end{equation}
For action loss, we use cross entropy loss, which is effective for classification task 
\begin{equation}
L_{act} = -\frac{1}{N}\Sigma_i^N\left(a_i\log{q_i^1} + (1-a_i)q_i^0 \right),
\end{equation}
where $a_i$ is the actioness label for $i_{th}$ sample, $q_i$ is a vector contains two elements which are predicted foreground and background score with Softmax activation. 
For localization loss, we adopt the widely used Smooth $L_1$ loss\cite{faster-rcnn}.
\begin{equation}
L_{loc} = \frac{1}{N_{pos}}\Sigma_i^{N_{pos}}\Sigma_{j=1}^2 smooth_{L1}(r_i^j-s_i^j)
\end{equation}
where $N_{pos}$ is the number of points we define as positive samples, $r_i$ is boundary prediction of $i_{th}$ point and $s_i$ is the target defined in Section \ref{sa}.

\section{Experiments}
\label{exper}
\subsection{Dataset and Setup}

\textbf{THUMOS 2014.}
We evaluate the proposed method on THUMOS 2014 dataset \cite{THUMOS14}, which is a standard and widely used dataset for temporal action proposal generation task. It contains 200 validation and 213 test untrimmed videos whose action instances are annotated temporally. Following the conventions \cite{ssn,s3d,turn,sst,bsn,mgg}, We train our models on validation set and evaluate them on testing set.

\textbf{Evaluation Metrics.}
For temporal action proposal generation, we adopt the conventional evaluation metric. We calculate Average Recall (AR) which is mean value of recall over different tIOU thresholds under various Average Number of proposals, denoted as AR@AN. Specifically, tIOU set of $[0.5:0.05:1.0]$ is used in our experiments. 

\textbf{Experiments Setup.}
During training, we used the stochastic gradient descent (SGD) as our optimizer. Momentum factor is set as 0.9 and weight decay factor is set as 0.0001 to regularize weights. We apply a multi-step learning scheduler to adjust learning rate. For all models, the training process lasts for 10 epochs. The initial learning rate is set as 0.0001. It is divided by 10 at epoch 7 and divided by 10 again at epoch 10. Training for one epoch means iterating over the dataset once. To form a batch while training, we clip videos as segments with equivalent length, which is 256 frames in our experiments specifically. The overlap of adjacent clips is 128 frames. We adopt sampling frequency of 8 fps in our experiments. According to our network architecture introduced in Section.\ref{arch_txt}, we finally get 126 samples for one clip regardless of assignment strategy.  To reduce overfitting, we adopt a multi-scale crop strategy\cite{wang2015towards} for per frame in addition to random horizontal flip transformation. Like most foreground/background tasks, huge imbalance of positive and negative samples exists in our experiments. Thus, we randomly sample negative samples in each batch to keep the ratio of positive and negative samples about 1:1. This strategy is proved to be efficient and results in more stable training.

During inference, We predict actioness score and boundary offset for each point in all feature maps. Final boundary can be computed by inverse transformation of Eq.(\ref{reg_target}). Then proposals of different clips in the same video are gathered. Finally, all proposals of a video are sorted according to the actioness score and filterd by Non-Maximum Suppression (NMS) with threshold value of 0.7. 

\subsection{Ablation Study}
\textbf{Comparison with pre-defined sliding windows.}
For sliding window based methods, the density of sliding windows at each timestamp is an important factor that influences the performance.
Most of them adopt a multi-scale anchor strategy to cover more ground truth proposals \cite{lin2017single,s3d}. It may come to an assumption that more dense pre-defined sliding windows will lead to a better result. To explore the influence of sliding window density, we setup a fair contrast experiment and results are shown in Table \ref{tab:slide-window}. For better comparison with our methods, we use the same architecture in Figure \ref{arch} and assign a base sliding window for each point in feature maps. The ratios in Table \ref{tab:slide-window} means the number of sliding windows in each point. For example, in second row, there are two pre-defined sliding windows for each position in feature maps. One is the base sliding window, the other is a sliding window with same center but half length as base sliding window. Thus, the amount of output proposals is twice as our method. During training for sliding window based methods, we assign positive labels for pre-defined sliding windows when their tIOU with any ground truth exceeds 0.5\cite{gao2017cascaded,lin2017single,s3d}. 

\begin{table}[h]
 \caption{Contrast of sliding windows with various ratios and DPP}
 \label{tab:slide-window}
 \centerline{
 \begin{tabular}{c|c|ccc}
    \hline
    method & ratios & AR@50 & AR@100 &AR@200\\
    \hline
    sliding window & 1 & 24.2 & 32.05 & 39.63 \\
    sliding window & 2 & 24.25 & 32.3 & 40.76 \\
    sliding window & 3 & 24.42 & 32.79 & 41.09 \\
    sliding window & 5 & 23.07 & 31.08 & 39.78 \\
    \hline
    dpp & n/a & \textbf{25.88} & \textbf{34.79} & \textbf{43.37} \\
 \hline
\end{tabular}
}
\end{table}

With a certain limit, more sliding windows do result in a higher average recall. However, over-density sliding windows do not help. While our method is superior to the best performance of sliding window based methods. This may be caused by many reasons. One possible reason is that multi-ratio sliding windows cause the ambiguous problem. Sliding windows at the same position with different ratios share the same input features, but expected to have different predictions. And our scale assignment strategy restricts target predictions of each point inside its receptive field, likely to result in better performance. Meanwhile, more sliding windows mean more outputs both in training and inference, undoubtedly leading to decrease in speed.  In conclusion, compared with sliding window based methods, DPP has the following advantages: (1) no ambiguous problem thus making optimization much easier; (2) fewer hyper-parameters which needs to be manually designed; (3) fewer proposal candidates resulting in faster processing.

\textbf{Analysis of Scale Assignment.} 
We design a novel scale assignment strategy in Section \ref{sa}. And according to Eq.(\ref{reg_target}), $\eta$ decides the localization target range of each pyramid. As $\eta$ increases, the localization target range will be larger. Thus a ground truth is more likely to match different pyramids, resulting in more positive proposal candidates.

\begin{table}[h]
 \caption{Influence of $\eta$ for DPP}
 \label{tab:eta}
 \centerline{
 \begin{tabular}{c|c|ccc}
    \hline
    $\eta$ & Backbone & AR@50 & AR@100 &AR@200\\
    \hline
     2 & ResNet-50 & 25.58 & 33.29 & 41.52 \\
     2.5 & ResNet-50 & 25.79 & 33.54 & 42.24 \\
     3 & ResNet-50 & \textbf{25.88} & \textbf{34.79} & \textbf{43.37} \\
     4 & ResNet-50 & 25.47 & 33.74 & 42.26 \\
 \hline
\end{tabular}
}
\end{table}

Table \ref{tab:eta} shows the influence of $\eta$ on the performance of DPP. And $\eta=3$ gets the best performance, which is used in all the following experiments. We can compute by the inverse transformation of Eq.(\ref{reg_target}) that, when $\eta=3$, the lower bound and upper bound of localization target are about $\frac{1}{3}$ and three times of default size for each pyramid.

\textbf{Exploration of Backbone Network.}
For the test of different backbones, we fix the pyramid amount as 6. As Table \ref{tab:backbone} shows, 3D ResNet-50, 3D ResNet-101\cite{hara2017learning} and C3D\cite{tran2015learning} are compared in our experiments. Backbone network with heavier weights tends to get better performances. We also test the performance of different backbone networks in speed. C3D outperforms other backbones in average recall but loses in speed competition. With almost the same average recall, 3D ResNet-101 attains about the twice speed of C3D. Note that all fps data is evaluated on a single GeForce GTX 1080 Ti. And for each experiment, fps is computed as mean fps of three epochs.

\begin{table}[h]
 \caption{Performance of different backbones}
 \label{tab:backbone}
 \centerline{
 \begin{tabular}{c|ccc|c}
    \hline
    Backbone Network & AR@50 & AR@100 &AR@200 & fps\\
    \hline
     ResNet-50 & 25.88 & 34.79 & 43.37 & 1804\\
     ResNet-101 & 28.01 & 36.27 & 44.36 & 1294 \\
     C3D & \textbf{28.57} & \textbf{36.65} & \textbf{44.55} & 676 \\
 \hline
\end{tabular}
}
\end{table}

\textbf{Varying Pyramids for DPP.} DPP adopts a pyramid structure to generate feature maps with different scales. We make a contrast experiment here to explore how pyramid amounts affect the performance of DPP. 

\begin{table}[h]
 \caption{Varying Pyramids for DPP}
 \label{tab:pyramid}
 \centerline{
 \begin{tabular}{c|c|ccc}
    \hline
    pyramids & npc & AR@50 & AR@100 &AR@200\\
    \hline
     6 & 126 & 28.57 & \textbf{36.65} & \textbf{44.55}\\
     5 & 124 & 27.14 & 35.61 & 43.51 \\
     4 & 120 & 27.28 & 35.69 & 42.89 \\
     3 & 112 & \textbf{28.75} & 36.22 & 43.05 \\
 \hline
\end{tabular}
}
\end{table}

Table \ref{tab:pyramid} shows results of different pyramid amounts varying from 3 to 6, where npc means number of proposals in one clip. Here, all experiments in Table \ref{tab:pyramid} use C3D as backbone network. It is found that under metrics of AR@100 and AR@200, 6 pyramids performs best. And under metirc of AR@50, 3 pyramids performs best. Since the difference among results of all these experiments is slight, we can infer that our proposed DPP is robust for pyramids variation.

\subsection{Comparison with State-of-the-art Methods}
We compare the proposed DPP with other state-of-the-art methods on action temporal proposal generation in Table \ref{tab:final}. To illustrate effectiveness of DPP, all methods adopt C3D \cite{tran2015learning} to extract spatio-temporal features and our method outperforms other methods. 
\begin{table*}[h]
 \caption{Comparison with  other temporal action proposal generation methods}
 \label{tab:final}
 \centerline{
 \begin{tabular}{p{4cm}|c|ccc|c}
    \hline
     & Features & AR@50 & AR@100 & AR@200 & fps \\
     \hline
     \noindent\emph{Sliding-window Methods}&&&&& \\
    \quad\quad DAPs\cite{escorcia2016daps} & C3D & 13.56 & 23.83 & 33.96 & 134.1 \\
    \quad\quad SCNN-prop\cite{shou2016temporal} & C3D & 17.22 & 26.17 & 37.01 & 60\\
    \quad\quad SST\cite{sst} & C3D & 19.90 & 28.36 & 37.90 & 308\\
    \quad\quad TURN\cite{turn} & C3D & 19.63 & 27.96 & 38.34 &880 \\
    \hline
    \noindent\emph{Actioness-grouping Methods}&&&&& \\
    \quad\quad BSN\cite{bsn} & C3D & 27.19 & 35.38 & 43.61 & - \\
    \hline
    \noindent\emph{Ensemble Methods}&&&&&\\
    \quad\quad MGG\cite{mgg} & C3D & 29.11 & 36.31 & 44.32 & - \\
    \hline
    \noindent\emph{Our Method}&&&&&\\
    \quad\quad DPP & C3D & 28.57 & 36.65 & 44.55 & 676\\
    \quad\quad DPP & ResNet-101 & 28.01 & 36.27 & 44.36 & 1294\\
 \hline
\end{tabular}
}
\end{table*}
All methods in the top part of the table adopt pre-defined sliding windows to generate proposal candidates, which is similar to anchor-based methods in object detection such as SSD\cite{ssd}. As we can see, DPP surpasses all sliding-window based method by a large margin. Specifically, DPP outperforms TURN, which performs best in sliding-window based methods, by improvement of  $16.2\%$ in AR@200. 

Actioness-grouping methods like BSN group temporal points with high actioness scores to form temporal action proposals. Compared to BSN, DPP increases AR@200 with $2\%$. MGG ensembles actioness-grouping based method which is proposed in \cite{ssn} and sliding-window based method to get higher results. Such methods cost much time when predicting, while our method generates high quality proposals with a high speed. Fps for the four methods in the top part of Table \ref{tab:final} are evaluated on a Geforce Titan X GPU and our method is evaluated on a Geforce GTX 1080 Ti GPU. Though BSN and MGG do not report their fps, according to the difference in principles, sliding-window based methods are expected to run faster than actioness-grouping based methods. Thus, compared to ensemble methods, DPP achieves comparative even better results with a much faster speed.

\section{Conclusion}
In this paper, We present a simple yet efficient method named Deep Point-wise Prediction to generate high quality temporal action proposals. Unlike previous work, we do not use any pre-defined sliding windows to generate proposal candidates, but predict left and right offsets for each point in different feature maps directly. We also note that there are also previous works in 2D object detection sharing similar ideas \cite{foveabox,densebox}. Without ambiguity of using same feature to regress different proposal candidates, our method gets better performance on localization and generates higher quality proposals.  In experiments, we explore different settings of our methods and prove its robustness. DPP is evaluated on standard THUMOS 2014 dataset to demonstrate its effectiveness. 

\bibliographystyle{splncs04}
\bibliography{sample}

\end{document}